\relax
\documentclass[letterpaper]{article} 
\usepackage{aaai22}  
\usepackage{times}  
\usepackage{helvet}  
\usepackage{courier}  
\usepackage[hyphens]{url}  
\usepackage{graphicx} 
\usepackage{natbib}  
\usepackage{caption} 
\DeclareCaptionStyle{ruled}{labelfont=normalfont,labelsep=colon,strut=off} 
\usepackage{multirow}
\usepackage{amsfonts}
\usepackage{booktabs}
\usepackage{algorithm}
\usepackage{algorithmic}
\usepackage[table,xcdraw]{xcolor}
\usepackage[colorlinks]{hyperref}
\hypersetup{
	colorlinks=true,
	linkcolor=red,
	filecolor=red,
	citecolor=black,      
	urlcolor=magenta,
}
\usepackage{newfloat}
\usepackage{listings}
\floatstyle{ruled}
\newfloat{listing}{tb}{lst}{}
\floatname{listing}{Listing}
\title{TransVOS: Video Object Segmentation with Transformers}
\author{
    Jianbiao Mei\thanks{Equal contribution}, Mengmeng Wang\footnotemark[1], Yeneng Lin, Yi Yuan, Yong Liu\thanks{Corresponding author}\\}

\affiliations{
    Laboratory of Advanced Perception on Robotics and Intelligent Learning,\\
	College of Control Science and Engineering,\\
	Zhejiang University; NetEase Fuxi AI Lab\\
	\texttt{\{jianbiaomei, mengmengwang, 3140100486\}}@zju.edu.cn, \\
	\texttt{yuanyi@corp.netease.com}, \texttt{yongliu@iipc.zju.edu.cn} \\
}

\usepackage{bibentry}

\begin{document}

\maketitle

\begin{abstract}
Recently, Space-Time Memory Network (STM) based methods have achieved state-of-the-art performance in semi-supervised video object segmentation (VOS). A crucial problem in this task is how to model the dependency both among different frames and inside every frame. However, most of these methods neglect the spatial relationships (inside each frame) and do not make full use of the temporal relationships (among different frames). In this paper, we propose a new transformer-based framework, termed TransVOS, introducing a vision transformer to fully exploit and model both the temporal and spatial relationships.
Moreover, most STM-based approaches employ two separate encoders to extract features of two significant inputs, i.e., reference sets (history frames with predicted masks) and query frame (current frame), respectively, increasing the models' parameters and complexity. To slim the popular two-encoder pipeline while keeping the effectiveness, we design a single two-path feature extractor to encode the above two inputs in a unified way. Extensive experiments demonstrate the superiority of our TransVOS over state-of-the-art methods on both DAVIS and YouTube-VOS datasets. Codes are available at \url{https://github.com/sallymmx/TransVOS.git}.
\end{abstract}

	\section{Introduction}
	Video Object Segmentation (VOS), as a fundamental task in the computer vision community, attracts more and more attention in recent years due to its potential application in video editing, autonomous driving, etc. In this paper, we focus on semi-supervised VOS, which provides the target objects' masks in the first frame, and the algorithms should produce the segmentation masks for those objects in the subsequent frames. Under this setting, VOS remains challenging due to object occlusion, deformation, appearance variation, and similar object confusion in video sequences.
	
	\begin{figure} [ht]
    \centering
	\includegraphics[width=\linewidth]{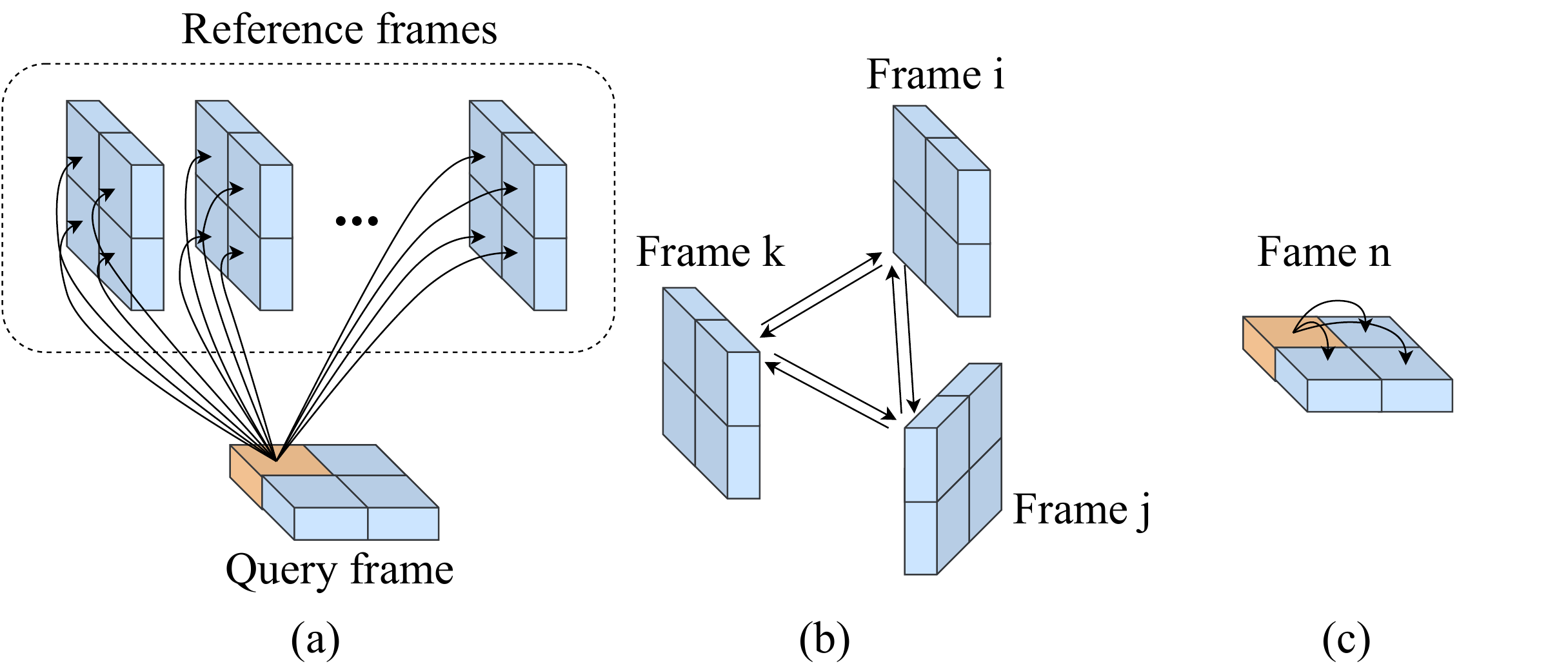}
	\caption{(a) relationships between pixels in query frame and pixels in reference frames. (b) relationships among pixels in different frames.  (c) relationships among pixels in the same frame.}
	\label{fig:relationships}
    \end{figure}
    
	A key problem in semi-supervised VOS is how to exploit the dependency both among different frames and inside every frame. To better depict this point, we define two relationships in this paper, i.e., \textbf{temporal relationships} (Fig. \ref{fig:relationships} (b)) and \textbf{spatial relationships} (Fig. \ref{fig:relationships} (c)). The former is the relationships among pixels in different frames, representing correspondence of target objects among all the frames, which is vital for learning robust global target object features and helps handle appearance change across frames. The latter is the relationships among pixels in one specific frame, including object appearance information for target localization and segmentation, which is important for learning local target object structure and helps obtain accurate masks. Recently, a group of matching-based methods \cite{hu2018videomatch, voigtlaender2019feelvos, yang2020collaborative, oh2019video, li2020fast, lu2020video, lai2020mast, liang2020video, seong2020kernelized, xie2021efficient, wang2021swiftnet} provide partial solutions for capturing above correspondence and achieve competitive performance. The basic idea of these methods is to compute the similarities of target objects between the current and past frames by feature matching, in which attention mechanism is widely used. Among them, the Space-Time Memory (STM) based approaches \cite{oh2019video, li2020fast, lu2020video, lai2020mast, liang2020video, xie2021efficient, wang2021swiftnet} have achieved great success. They propose to apply spatio-temporal attention between every pixel in previous frames and every pixel in the current frame. 
	However, the spatio-temporal attention module in previous approaches could not thoroughly model the temporal and spatial relationships. As illustrated in Fig. \ref{fig:relationships} (a), it only computes attentions among pixels in the query frame against pixels in each reference frame. The temporal relationships are not fully explored since they ignore the dependency among all the history frames. Besides, the spatial relationships of pixels inside every frame are also neglected. There are a few methods paid attention to these issues. For instance, EGMN \cite{lu2020video} proposes a fully-connected graph to capture cross-frame correlation, which exploits the temporal relationships effectively. However, EGMN still omits the spatial relationships. Our key insight is that both the temporal and spatial relationships are significant to VOS and should be utilized effectively. We find the recent vision transformer \cite{dosovitskiy2020image} is a satisfactory model to cater to the demand since the self-attention modules in the transformer could establish dependency among every elements in the input sequence. Therefore, we proposed a new transformer-based architecture for VOS, termed TransVOS, to capture the temporal relationships and spatial relationships jointly.
	
	Another significant but open problem is how to effectively represent the inputs, i.e., reference sets (history information) and query frames (current information). Many existing methods try to encode these two kinds of information with two separate encoders (always termed as memory/reference encoder and query encoder), since reference sets include several history images and their predicted masks while the query frames only has current images. This two-encoder pipeline is swollen and contains plenty of redundant parameters. Existing ways to slim it are limited. For example, AGAME \cite{johnander2019generative} and RANet \cite{wang2019ranet} employ a siamese network to encode reference and query frame and then concatenate features of the reference frame with its predicted mask to reinforce target features. RGMP \cite{oh2018fast} and AGSS-VOS \cite{lin2019agss} concatenate the current frame with the previous frame's mask or warped mask to form a 4-channel input, so as the reference sets. Then a shared encoder with a 4-channel input layer is used to extract features. These strategies can reuse features and effectively reduce the amount of parameters. Nevertheless, directly concatenating them with high-level semantic features is insufficient since the abundant features like contour and edges of the mask are not fully leveraged. Besides, due to the padding operation, object positions of frame features and the down sampled mask may be misaligned. In addition, concatenating the previous frame's mask with the query frame may bring large displacement shifting error and using optical flow to warp the mask is time-consuming. Different from them, in this paper, we develop a plain yet effective feature extractor which has a two-path input layer and accepts the reference sets and the query frames in the meanwhile, significantly simplifying the existing VOS framework while keeping the effectiveness. 
	
	Our main contributions can be summarized as follows:
	
	$\bullet$ We proposed a new transformer-based VOS framework, TransVOS, to effectively model the temporal and spatial dependency. 
	
    $\bullet$ TransVOS has a concise pipeline, which does not need multiple encoders, dramatically simplifying the existing VOS pipelines.
	
	$\bullet$ We comprehensively evaluate the proposed TransVOS on 3 benchmark datasets including DAVIS 2016/2017 \cite{perazzi2016benchmark, pont20172017} and YouTube-VOS \cite{xu2018youtube}. The results demonstrate the effectiveness and efficiency of our method in comparison with the previous approaches.
	
	\section{Related works}
	\paragraph{Tracking-based methods.} These methods \cite{wang2019fast,  chen2020state, voigtlaender2020siam, huang2020fast} integrate object tracking techniques to indicate target location and spatial area for segmentation and improve the inference speed. SiamMask \cite{wang2019fast} adds a mask branch on SiamRPN \cite{li2018high} to narrow the gap between tracking and segmentation. FTAN-DTM \cite{huang2020fast} takes object segmentation as a sub-task of tracking, introducing “tracking-by-detection” model into VOS. While the accuracy of tracking often limits these methods' performance.
	SAT \cite{chen2020state} and FTMU \cite{sun2020fast} fuse object tracking and segmentation into a truly unified pipeline. SAT combines SiamFC++ \cite{xu2020siamfc++} and proposed an estimation-feedback mechanism to switch between mask box and tracking box, making segmentation and tracking tasks enhance each other.  
	\paragraph{Matching-based methods.} Recently, state-of-the-art performance has been achieved by matching-based methods \cite{hu2018videomatch, voigtlaender2019feelvos, wang2021swiftnet, oh2019video, lu2020video, liang2020video, xie2021efficient}, which perform feature matching to learn target object appearances
	offline. FEELVOS \cite{voigtlaender2019feelvos} and CFBI \cite{yang2020collaborative} perform the nearest neighbor matching between the current frame and the first and previous frames in the feature space. STM \cite{oh2019video} introduces an external memory to store past frames' features and uses the attention-based matching method to retrieve information from memory. KMN \cite{seong2020kernelized} proposes to employ a gaussian kernel to reduce the non-locality of the STM. RMNet \cite{xie2021efficient} proposes to replace STM's global-to-global matching with local-to-local matching to alleviate the ambiguity of similar objects. EGMN \cite{lu2020video} organizes the memory network as a fully connected graph to store frames as nodes and capture cross-frame relationships by edges. However, these methods do not fully utilize the spatio-temporal relationships among reference sets and query frame. In this paper, we introduce a vision transformer to model spatio-temporal dependency, which can help handle large object appearance change.
	\paragraph{Transformer-based methods.} 
	Recently, transformer has achieved great success in vision tasks like image classification \cite{dosovitskiy2020image}, object detection \cite{carion2020end}, semantic segmentation \cite{wang2020max}, object tracking \cite{yan2021learning}, etc.
	Due to the importance of spatial and temporal relationships for segmenting, we also employ the vision transformer into the VOS task, which is inspired by DETR \cite{carion2020end}. 
    We find two transformer-based methods, SST \cite{duke2021sstvos} and JOINT \cite{mao2021joint}. SST uses the transformer's encoder with sparse attention to capture the spatio-temporal information among the current frame and its preceding frames. While mask representations are not explored in SST. JOINT combines inductive and transductive learning and extend the transduction branch to a transformer architecture. But its network structure is complicated.
    Besides, both SST and JOINT did not employ the transformer's decoder and could not enjoy the strong power of it.

	
	
	\section{Methods}
	The overview of our framework is illustrated in Fig.  \ref{fig:pipeline}. It mainly consists of a feature extractor , a vision transformer, a target attention block and a segmentation head. When segmenting a specific frame, we firstly use the feature extractor to extract the features of the current frame and reference sets. The outputs of the extractor are fed into a bottleneck layer to reduce the channel number. Then features are flattened before feeding into a vision transformer, which models the temporal and spatial relationships. Moreover, the target attention block takes both the transformer's encoder and decoder's outputs as input and then outputs the feature maps, representing the target mask features. Finally, a segmentation head is attached after the target attention block to obtain the predicted object mask.
	
	\subsection{Feature Embedding} \label{sec2_1}
\begin{figure*} [ht]
		\centering
		\includegraphics[width=\linewidth]{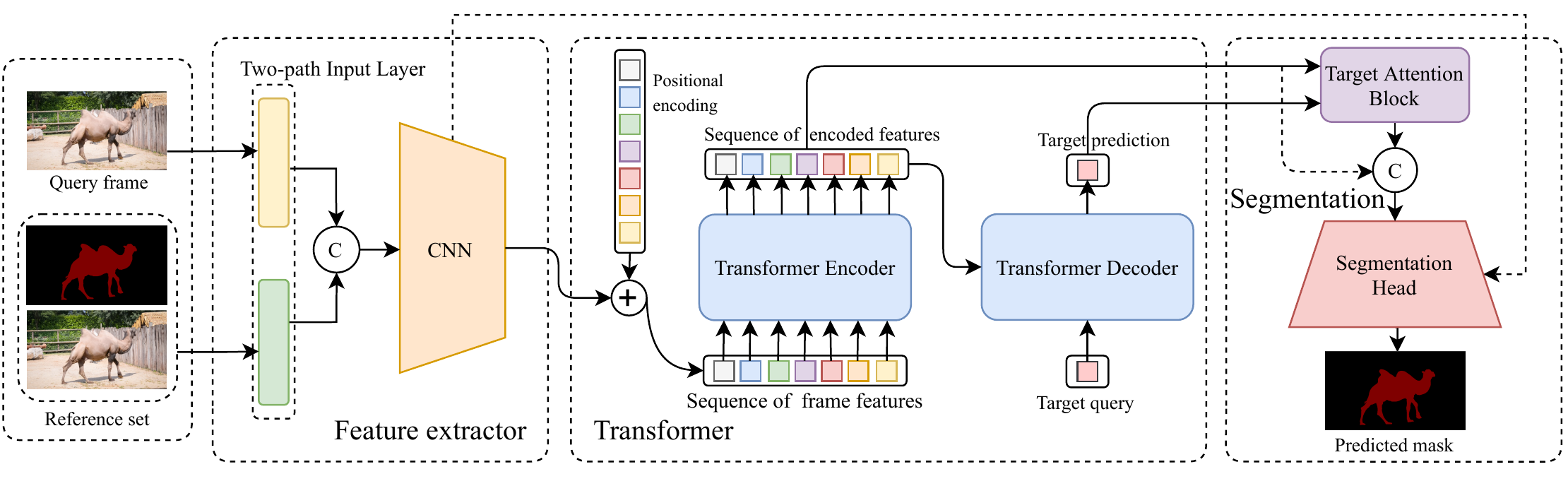}
		\caption{Overview of our TransVOS. The feature extractor is used to extract the features of the current frame and reference sets. The vision transformer is exploited to model the temporal and spatial relationships. The target attention block (TAB) is used to extract the target mask features. The segmentation head is designed to obtain the predicted object mask. "$+$", "$\rm C$" indicate adding and concatenating operation, respectively.}
		\label{fig:pipeline}
	\end{figure*}

	To fully exploit the temporal and spatial information in the reference sets and the query frame, we need a delicate feature extractor that can effectively extract the target object features and map them into an embedding space to be ready for feeding into the following vision transformer.
	\paragraph{Feature extractor.} TransVOS uses a single feature extractor to extract features of the query frame and reference sets in a unified way. Specifically, we design a two-path input layer to adaptively encode two types of inputs, i.e, RGB frames or the pairs of RGB frames with corresponding object masks. As shown in Fig. \ref{fig:pipeline}, when taking the RGB frames as input, it will go through the first path which has one regular convolution operation. For reference sets, the second path will be used, which contains three convolutions to encode the RGB frame, the object mask's foreground and background, respectively. The output features of the three convolutions are added together to output the features. Our method can use arbitrary convolutional networks as the feature extractor by replacing the first layer with the two-path input layer. Here we employ the first four stages of ResNet \cite{he2016deep} as the feature extractor. After going through the two-path input layer, the features from the query frame and reference sets are first concatenated along the temporal dimension and then fed into the convolutional network (CNN). Finally, the reference sets and current frame are mapped to feature maps $\textbf{f}\in\mathbb{R}^{(T+1) \times C \times H \times W}$, where $H$, $W$, $C$ are the height, width and channel. $T$ is the number of the reference pairs. 
	The proposed two-path feature extractor has much fewer parameters (about 20\% reduction) than the traditional two-encoders pipeline while keeping the effectiveness.
	
	Before feeding into the vision transformer, we use a 1x1 convolution layer to reduce the spatial channel of the feature maps from $C$ to $d$ $(d<C)$, resulting in new feature maps $\textbf{f}_1\in\mathbb{R}^{(T+1) \times d \times H \times W}$. Then, the spatial and temporal dimensions of $\textbf{f}_1$ are flattened into one dimension, producing feature vectors $\textbf{X}\in \mathbb{R} ^{(T+1)HW \times d}$, which servers as the input of the transformer encoder.
	
	\subsection{Relationship Modeling} \label{sec2.2}
	Transformers have strong capabilities for modeling spatio-temporal relationships. First, the positional encoding explicitly introduces space-time position information, which could help the encoder model spatio-temporal relationships among pixels in the input frames. Second, the encoder could learn the target object's correspondence among the input frames and model the target object's structure in a specific frame. Third, the decoder could predict the spatial positions of the target objects in the query frame and focus on the most relevant object, which learns robust target representations for target object and empowers our network to handle similar object confusion better.
	\paragraph{Positional encoding.}
	We add sinusoidal positional encoding $PE$ \cite{vaswani2017attention} to the embedded features $\textbf{X}$ to form the inputs $\textbf{Z}$ of the transformer. Mathematically,
	\begin{equation}
		\textbf{Z} = \textbf{X} + PE 
	\end{equation}
	\begin{equation}
		PE(pos, 2i) = sin(pos/10000^{2i/d})
	\end{equation}
	\begin{equation}
		PE(pos, 2i+1) = cos(pos/10000^{2i/d})
	\end{equation} where $pos$ and $i$ are the spatio-temporal position and the dimension of the features $\textbf{X}$, respectively.
	
	\paragraph{Transformer encoder.} 
	The transformer encoder is used to model the spatio-temporal relationships among pixel-level features 
	of the sequence. It takes features $\mathbf{Z}$ as input and outputs encoded features $\textbf{E}$.
	The encoder consists of $N$ encoder layers, each of which has a standard architecture including a multi-head self-attention module and a fully connected feed-forward network.
	The multi-head self-attention module is used to capture spatio-temporal relationships from different representation sub-spaces. Let $\mathbf{z}_{p, t} \in \mathbb{R}^{d}$ represents an element of $\mathbf{Z}$, where $p$ and $t$ denote the spatial and temporal position, respectively. Firstly, for $m_E$-th $(m_E \leq M)$ attention head, the query/key/value vector ($\mathbf{q}^{m_E}_{p, t} / \mathbf{k}^{m_E}_{p, t} / \mathbf{v}^{m_E}_{p, t}$) is computed from the representation $\mathbf{z}_{p, t}$:
	\begin{equation}
		\mathbf{q}^{m_E}_{p, t} = \mathbf{W}^{m_E}_{q}\mathbf{z}_{p, t},
		\mathbf{k}^{m_E}_{p, t} = \mathbf{W}^{m_E}_{k}\mathbf{z}_{p, t},
		\mathbf{v}^{m_E}_{p, t} = \mathbf{W}^{m_E}_{v}\mathbf{z}_{p, t}
	\end{equation}
	Then the multi-head self-attention feature $\mathbf{\hat{z}}_{p, t}$ is calculated by
	\begin{equation}
		\mathbf{\hat{z}}_{p, t} = \sum_{m_E=1}^{M}\mathbf{W}_{o}^{m_E}[\sum_{t=1}^{T+1}\sum_{p=1}^{HW}\sigma(\frac{(\mathbf{q}^{m_E}_{p, t})^{T}\mathbf{k}^{m_E}_{p, t}}{\sqrt{d_{m_E}}})\cdot \mathbf{v}^{m_E}_{p, t}]
	\end{equation}
	where $T$ represents the size of the referent set, $\mathbf{W}_{q}^{m_E}, \mathbf{W}_{k}^{m_E}, \mathbf{W}_{v}^{m_E}\in \mathbb{R}^{d_{m_E}\times d}$ and $\mathbf{W}_{o}^{m_E} \in \mathbb{R}^{d\times d_{m_E}}$ are learnable weights ($d_{m_E}=d/M$ by default), $\sigma$ indicates the $softmax$ function. Note that we compute attention along the spatio-temporal dimension. Thus we can model the spatial relationships and temporal relationships in the meanwhile.
	\paragraph{Transformer decoder.}
	The transformer decoder aims to predict the spatial positions of the target and focus on the most relevant object in the query frame. It takes encoded features $\mathbf{E}$ and target query $\mathbf{x}_{o}$ as input and output decoded features $\mathbf{O}$. We only utilize one target query in the decoder to query the target object features since there is only one prediction. The decoder consists of $N$ decoder layers, each of them includes a multi-head self-attention module, a multi-head cross-attention module, and a fully connected feed-forward network. In our TransVOS, the multi-head self-attention module is used to integrate target information from different representation sub-space. And the multi-head cross-attention module is mainly leveraged to fuse target object features from the encoder. With only one target query $\mathbf{x}_{o} \in \mathbb{R}^{d}$, the multi-head self-attention feature $\mathbf{\hat x}_{so}$ can be expressed as:
	\begin{equation}
	    \mathbf{\hat x}_{so} = \sum_{m_D=1}^{M}\mathbf{W}_{o}^{m_D}(\mathbf{W}_{v}^{m_D}\mathbf{x}_{o})
	\end{equation} where $m_D$ indexes the attention head in multi-head self-attention module, $\mathbf{W}_{v}^{m_D}\in \mathbb{R}^{d_{m_D}\times d}$ and $\mathbf{W}_{o}^{m_D} \in \mathbb{R}^{d\times d_{m_D}}$ are learnable weights ($d_{m_D}=d/M$ by default).
	
	For the multi-head cross-attention module, let $\mathbf{e}_{p, t} \in \mathbb{R}^{d}$ represents an element of $\mathbf{E}$, $p$ and $t$ denote the spatial and temporal position, respectively. For $m'_D$-th ($m'_D \leq M$) attention head, the key and value vectors $\mathbf{k}^{m'_D}_{p, t}, \mathbf{v}^{m'_D}_{p, t}$ are computed as:
	\begin{equation}
		\mathbf{k}^{m'_D}_{p, t} = \mathbf{W}^{m'_D}_{k}\mathbf{e}_{p, t},
		\mathbf{v}^{m'_D}_{p, t} = \mathbf{W}^{m'_D}_{v}\mathbf{e}_{p, t}
	\end{equation}
	Then with the query $\mathbf{x'}_{o} \in \mathbb{R}^{d}$, the cross-attention feature $\mathbf{\hat x}_{co}$ is calculated:
	\begin{equation}
		\mathbf{\hat x}_{co} = \sum_{m'_D=1}^{M}\mathbf{W}_{o}^{m'_D}[\sum_{t=1}^{T+1}\sum_{p=1}^{HW}\sigma(\frac{(\mathbf{W}_{q}^{m'_D}\mathbf{x'}_{o})^{T}\mathbf{k}^{m'_D}_{p, t}}{\sqrt{d_{m'_D}}})\cdot \mathbf{v}^{m'_D}_{p, t}]
	\end{equation} where $T$ denotes the size of the reference set, $\mathbf{W}_{q}^{m'_D}, \mathbf{W}_{k}^{m'_D}, \mathbf{W}_{v}^{m'_D}\in \mathbb{R}^{d_{m'_D}\times d}$ and $\mathbf{W}_{o}^{m'_D} \in \mathbb{R}^{d\times d_{m'_D}}$ are learnable weights($d_{m'_D}=d/M$ by default). $\sigma$ indicates the $softmax$ function.
	
	\subsection{Segmentation} \label{sec2.3}
	\paragraph{Target attention block.} To obtain the target mask prediction from the outputs of the transformer, the model needs to extract the target's mask features of the query frame. To achieve this goal, we use a Target Attention Block (TAB) to get an attention map first. TAB computes the similarity between encoded features $\mathbf{E}_Q$ of query frame and the output features $\mathbf{O}$ from the decoder. $\mathbf{O}$ and $\mathbf{E}_Q$ are fed into a multi-head self-attention module (with $M$ head) to obtain the attention maps. We concatenate the attention maps with $\mathbf{E}_Q$ as the input $\mathbf{S}$ of the following segmentation head to enhance the target features. The above procedure can be formulated as:
    \begin{equation}
	    Attn_{i}(\mathbf{O}, \mathbf{E}_Q) = \sigma(\frac{(\mathbf{W}_{q}^{i}\mathbf{O})^{T}(\mathbf{W}_{k}^{i}\mathbf{E}_Q)}{\sqrt{d_{i}}})
	\end{equation}
	\begin{equation}
	    \mathbf{S} = [\mathbf{E}_Q, Attn_{1}(\mathbf{O}, \mathbf{E}_Q), \cdots, Attn_{M}(\mathbf{O}, \mathbf{E}_Q)]
	\end{equation} where $i$ indexes the attention head,  $\mathbf{W}_{q}^{i}, \mathbf{W}_{k}^{i}\in \mathbb{R}^{d_{i}\times d}$, are learnable weights ($d_{i}=d/M$ by default).
	\paragraph{Segmentation head.}
	The features $\mathbf{S}$ are fed into a segmentation head which outputs the final mask prediction. Here, we use the refine module used in \cite{oh2018fast, oh2019video} as the building block to construct our segmentation head. It consists of two blocks, each of which takes both the output of the previous stage and the current frame's feature maps from feature extractor at the corresponding scale through skip-connections. Gradually upscale the compressed feature maps by a factor of two at a time. And a 2-channel convolution and a $softmax$ operation are attached behind blocks to attain the predicted mask in 1/4 scale of the input image. Finally, we use bi-linear interpolation to upscale the predicted mask to the original scale.
    \paragraph{Multi-object segmentation.}
	Our framework can be extended to multi-object segmentation easily. Specifically, the network first predicts the mask for every target object. Then, a soft aggregation operation is used to merge all the predicted maps. We apply this way during both the training and inference to keep consistency on both stages. For each location $l$ in predicted mask $\mathbf{M}_{i}$ of object $i(i<N)$, the probability $p_{l, i}$ after soft aggression operation can be expressed as:
	\begin{equation}
		p_{l, i} = \sigma({\rm logit}(\hat{p}_{l, i}))
		= \frac{\hat{p}_{l, i}/(1 - \hat{p}_{l, i})}{\sum_{j=0}^{N-1}{\hat{p}_{l, j}/(1 - \hat{p}_{l, j})}}
	\end{equation} where $N$ is the number of objects. $\sigma$ and logit represent the $softmax$ and $logit$ functions, respectively. The probability of the background is obtained by subtracting from 1 after computing the output of the merged foreground.
	
	\subsection{Training and Inference}
    \paragraph{Training.} Our proposed TransVOS doesn't require extremely long training video clips since it does not have any temporal smoothness assumptions. Even though, TransVOS can still learn long-term dependency. Just like most STM-based methods \cite{oh2019video, li2020fast, lu2020video, liang2020video, seong2020kernelized}, we synthesis video clips by applying data augmentations (random affine, color, flip, resize and crop) on a static image of datasets \cite{cheng2014global, lin2014microsoft, li2014secrets, everingham2010pascal}. Then we use the synthetic videos to pretrain our model. This pre-training procedure helps our model to be robust against a variety of object appearance and categories.
    After that, we train our model on real videos. We randomly select $T$ frames from a video sequence of DAVIS \cite{perazzi2016benchmark, pont20172017} or YouTube-VOS \cite{xu2018youtube} and apply data augmentation on those frames to form a training video clip. By doing so, we can expect our model to learn long-range spatio-temporal information.
	We add cross-entropy loss $\mathcal{L}_{cls}$ and mask IoU loss $\mathcal{L}_{IoU}$ as the multi-object training loss $\mathcal{L}$, which can be expressed as:
	\begin{equation}
		\mathcal{L} = \frac{1}{N} \sum_{i=0}^{N-1}[\mathcal{L}_{cls}(\mathbf{M}_{i}, \mathbf{Y}_{i}) + \mathcal{L}_{IoU}(\mathbf{M}_{i}, \mathbf{Y}_{i})]
	\end{equation}
	\begin{equation}
		\mathcal{L}_{cls}(\mathbf{M}_{i}, \mathbf{Y}_{i}) = -\frac{1}{|\Omega|}\sum_{p\in\Omega}[\mathbf{Y}_{i}
		{\rm log}(\frac{\exp(\mathbf{M}_{i})}
		{\sum_{j=0}^{N-1}(\exp(\mathbf{M}_{j}))})]_{p}
	\end{equation}
	\begin{equation}
		\mathcal{L}_{IoU}(\mathbf{M}_{i}, \mathbf{Y}_{i}) = 1 - \frac{{\sum_{p \in \Omega } {\min (\textbf{Y}^{p}_{i},\textbf{M}^{p}_{i})} }}{{\sum_{p \in \Omega } {\max (\textbf{Y}^{p}_{i},\textbf{M}^{p}_{i})}}}
	\end{equation} where $\Omega$ denotes the set of all pixels in the object mask, $\mathbf{M}_{i}, \mathbf{Y}_{i}$ represent the predicted mask and ground truth of object $i$, $N$ is the number of objects. Note that $N$ is set to 1 when segmenting a single object.
	
	\paragraph{Inference.} Our model uses past frames with corresponding predicted masks to segment the current frame during the online reference phase. We don't use external memory to store every past frame's features but only use the first frame with ground truth and the previous frames with its predicted masks as the reference sets. Because the former always provides the most reliable information, and the later is the most similar one to the current frame.

	\begin{table*}[t]
		\centering
		\resizebox{\textwidth}{!}{
			\begin{tabular}{lccccccccc}
				\cmidrule(r){1-9}
				\multicolumn{1}{c}{\multirow{2}{*}{Methods}} &  
				\multicolumn{1}{c}{\multirow{2}{*}{OL}} & 
				\multicolumn{4}{c}{DAVIS16 val} & 
				\multicolumn{3}{c}{DAVIS17 val} \\ 
				\cmidrule(r){3-6}  \cmidrule(r){7-9}
				\multicolumn{1}{c}{} & 
				\multicolumn{1}{c}{} & 
				\multicolumn{1}{c}{ FPS} & 
				\multicolumn{1}{c}{$\mathcal{J}\&\mathcal{F}(\%)$} & 
				\multicolumn{1}{c}{$\mathcal{J}(\%)$} & 
				\multicolumn{1}{c}{$\mathcal{F}(\%)$} & 
				\multicolumn{1}{c}{$\mathcal{J}\&\mathcal{F}(\%)$} & 
				\multicolumn{1}{c}{$\mathcal{J}(\%)$} & 
				\multicolumn{1}{c}{$\mathcal{F}(\%)$} \\ \cmidrule(r){1-9}
				OSVOS \cite{caelles2017one} & $\checkmark$ & 0.22 & 80.2 & 79.8 & 80.6 & 60.3 & 56.7 & 63.9 \\
				OnAVOS \cite{voigtlaender2017online} & $\checkmark$ & 0.08 & 85.5 & 86.1 & 84.9 & 67.9 & 64.5 & 71.2 \\
				PReMVOS \cite{luiten2018premvos} & $\checkmark$ & 0.03 & 86.8 & 84.9 & 88.6 & 77.8 & 73.9 & 81.7 \\
				FRTM(+YT) \cite{robinson2020learning} & $\checkmark$ & 21.9 & 83.5 & - & - & 76.7 & - & - \\ \cmidrule(r){1-9}
				RGMP \cite{oh2018fast} &  & 7.7 & 81.8 & 81.5 & 82.0 & 66.7 & 64.8 & 68.6 \\
				RaNet \cite{wang2019ranet} &  & \textbf{30} & 85.5 & 85.5 & 85.4 & 65.7 & 63.2 & 68.2 \\
				AFB-URR \cite{liang2020video} &  & - & - & - & -  & 74.6 & 73.0 & 76.1 \\
				AGAME(+YT) \cite{johnander2019generative} &  & 14.3 & 82.1 & 82.0 & 82.2 & 70.0 & 67.2 & 72.7 \\
				FEELVOS(+YT) \cite{voigtlaender2019feelvos} &  & 2.2 & 81.7 & 81.1 & 82.2 & 71.5 & 69.1 & 74.0 \\
				STM(+YT) \cite{oh2019video} &  & 6.3 & 89.3 & 88.7 & 89.9 & 81.8 & 79.2 & 84.3 \\
				KMN(+YT) \cite{seong2020kernelized} &  & 8.3 & \textbf{90.5} & 89.5 & \textbf{91.50} & 82.8 & 80.0 & 85.6 \\
				EGMN(+YT) \cite{lu2020video} &  & - & - & - & - & 82.8 & 80.2 & 85.2 \\
				CFBI(+YT) \cite{yang2020collaborative} &  & 6 & 89.4 & 88.3 & 90.5 & 81.9 & 79.1 & 84.6 \\
				JOINT(+YT) \cite{mao2021joint} &  & - & - & - & - & 83.5 & 80.8 & 86.2 \\
				SST(+YT) \cite{duke2021sstvos} &  & - & - & - & - & 82.5 & 79.9 & 85.1 \\
			    RMNet(+YT) \cite{xie2021efficient} &  & 12 & 88.8 & 88.9 & 88.7 & 83.5 & 81.0 & 86.0 \\
			    SwiftNet(+YT) \cite{wang2021swiftnet} &  & 25 & 90.4 & \textbf{90.5} & 90.3 & 81.1 & 78.3 & 83.9 \\
			    \cmidrule(r){1-9}
				\textbf{TransVOS} &  & 6.6 & 85.8 & 85.4 & 86.2 & 78.1 & 75.7 & 80.5 \\
				\textbf{TransVOS(+YT)} &  & 6.6 & \textbf{90.5} & 89.8 & 91.2 & \textbf{83.9} & \textbf{81.4} & \textbf{86.4} \\ \cmidrule(r){1-9}
		\end{tabular}}
		\caption{Comparison with state-of-the-art on the DAVIS16 and DAVIS17 validation set. `OL' indicates the use of online-learning strategy. `+YT' indicates the use of YouTube-VOS for training. Note that the runtime of other methods was obtained from the corresponding papers. }
		\label{table:1}
	\end{table*}
	
	\begin{table*}[t]
		\centering
		\resizebox{\textwidth}{!}{
			\begin{tabular}{lccccccccc}
				\cmidrule(r){1-10}
				\multicolumn{1}{c}{\multirow{2}{*}{Methods}} &
				\multicolumn{1}{c}{\multirow{2}{*}{OL}} &
				\multicolumn{3}{c}{DAVIS17 test-dev} & 
				\multicolumn{5}{c}{YouTube-VOS 2018 val} \\ \cmidrule(r){3-5} \cmidrule(r){6-10}
				\multicolumn{1}{c}{} & 
				\multicolumn{1}{c}{} & 
				\multicolumn{1}{c}{$\mathcal{J}\&\mathcal{F}(\%)$} & 
				\multicolumn{1}{c}{$\mathcal{J}(\%)$} & 
				\multicolumn{1}{c}{$\mathcal{F}(\%)$} & 
				\multicolumn{1}{c}{Overall} & 
				\multicolumn{1}{c}{$\mathcal{J}_{s}(\%)$} & 
				\multicolumn{1}{c}{$\mathcal{J}_{u}(\%)$} & 
				\multicolumn{1}{c}{$\mathcal{F}_{s}(\%)$} & 
				\multicolumn{1}{c}{$\mathcal{F}_{u}(\%)$} \\ \cmidrule(r){1-10}
				OSVOS \cite{caelles2017one} & $\checkmark$ & 50.9 & 47.0 & 54.8 & 58.8 & 59.8 & 54.2 & 60.5 & 60.7 \\
				OnAVOS \cite{voigtlaender2017online} & $\checkmark$ & 52.8 & 49.9 & 55.7 & 55.2 & 60.1 & 46.6 & 62.7 & 51.4 \\
				PReMVOS \cite{luiten2018premvos} & $\checkmark$ & 71.6 & 67.5 & 75.7 & - & - & - & - & - \\
				STM-cycle \cite{li2020delving} & $\checkmark$ & 58.6 & 55.3 & 62.0 & 70.8 & 72.2 & 62.8 & 76.3 & 71.9 \\
				\cmidrule(r){1-10}
				RGMP \cite{oh2018fast} &  & 52.9 & 51.3 & 54.4 & 53.8 & 59.5 & 45.2 & - & - \\
				AGAME \cite{johnander2019generative} &  & - & - & - & 66.1 & 67.8 & 60.8 & - & - \\
				FEELVOS \cite{voigtlaender2019feelvos} &  & 57.8 & 55.2 & 60.5 & - & - & - & - & - \\
				RaNet \cite{wang2019ranet} &  & 55.3 & 53.4 & - & - & - & - & - & - \\
				STM \cite{oh2019video} &  & 72.2 & 69.3 & 75.2 & 79.4 & 79.7 & 72.8 & 84.2 & 80.9 \\
				CFBI \cite{yang2020collaborative} &  & 74.8 & 71.1 & 78.5 & 81.4 & 81.1 & 75.3 & 85.8 & \textbf{83.4} \\
				AFB-URR \cite{liang2020video} & & - & - & - & 79.6 & 78.8 & 74.1 & 83.1 & 82.6 \\
				KMN \cite{seong2020kernelized} &  & \textbf{77.2} & \textbf{74.1} & 80.3 & 81.4 & 81.4 & 75.3 & 85.6 & 83.3 \\
				EGMN \cite{lu2020video} &  & - & - & - & 80.2 & 80.7 & 74.0 & 85.1 & 80.9 \\
				SST \cite{duke2021sstvos} &  & - & - & - & 81.7 & 81.2 & \textbf{76.0} & - & - \\
				RMNet \cite{xie2021efficient} &  & 75.0 & 71.9 & 78.1 & 81.5 & 82.1 & 75.7 & 85.7 & 82.4 \\
				SwiftNet \cite{wang2021swiftnet} &  & - & - & - & 77.8 & 77.8 & 72.3 & 81.8 & 79.5 \\
				\cmidrule(r){1-10}
				\textbf{TransVOS} & & 76.9 & 73.0 & \textbf{80.9} & \textbf{81.8} & 82.0 & 75.0 & \textbf{86.7} & \textbf{83.4} \\ \cmidrule(r){1-10}
		\end{tabular}}
		\caption{Compare to the state of the art on the DAVIS17 test-dev set and YouTube-VOS 2018 validation set. `OL' indicates the use of online-learning strategy. The subscripts of $\mathcal{J}$ and $\mathcal{F}$ on YouTube-VOS denote seen objects ($s$) and unseen objects ($u$). The metric overall means the average of $\mathcal{J}_s, \mathcal{J}_u, \mathcal{F}_s, \mathcal{F}_u$.}
		\label{table:2}
	\end{table*}
		
	\section{Experiments}
	We evaluate our approach on DAVIS \cite{perazzi2016benchmark, pont20172017} and YouTube-VOS \cite{xu2018youtube} benchmarks. Please refer to supplementary materials for more details about the datasets and evaluation metrics.
	\begin{table*} [ht]
		\centering
		\resizebox{\textwidth}{!}{
			\begin{tabular}{ccccccccccc}
				\cmidrule(r){1-11}
				{Variants} & {Mask utilization} & {Reference sets} & {$\mathcal{J}_{M}(\%)$} & {$\mathcal{J}_{R}(\%)$} & {$\mathcal{J}_{D}(\%)$} & {$\mathcal{F}_{M}(\%)$} & {$\mathcal{F}_{R}(\%)$} & {$\mathcal{F}_{D}(\%)$} & {$\mathcal{J\&F}(\%)$} & FPS \\ \cmidrule(r){1-11}
				1 & \textit{multiply} & 1st frame & 51.4 & 59.9 & 13.4 & 58.3 & 63.6 & 14.4 & 54.9 & - \\
				2 & \textit{residual} & 1st frame & 58.5 & 67.1 & 17.6 & 65.5 & 75.0 & 18.6 & 62.0 & - \\
				3 & \textit{two-path} & 1st frame & 66.5 & 78.2 & 13.3 & 73.6 & 83.5 & 15.9 & 70.0 & 23.0\\
				4 & \textit{two-path} & previous frame & 64.3 & 74.8 & 11.7 & 70.5 & 81.3 & 14 & 67.4 & 17.6\\
				5 & \textit{two-path} & 1st \& previous frames & 73.1 & 86.6 & 1.8 & 79.7 & 91.5 & 5.3 & 76.4 & 17.1\\
				6 & \textit{two-path} & Every 5 frames & 70.2 & 82.2 & 6.0 & 77.6 & 89.0 & 8.1 & 73.9 & 5.1\\
				\cmidrule(r){1-11}
		\end{tabular}}
		\caption{Ablation studies of mask utilization and reference sets with input resolution 240p on DAVIS 2017 validation set.}
		\label{AB:1}
	\end{table*}
	\subsection{Implementation details.} We use the first four stages of ResNet50 \cite{he2016deep} and replace its input layer with the proposed two-path input layer to form our feature extractor. The number of transformer encoder layers and decoder layers is set to $N=6$. The multi-head attention layers have $M = 8$ heads, width $d = 256$, while the feed-forward networks have hidden units of 2048. Dropout ratio of 0.1 is used. The TransVOS is trained with the input resolution of 480p, and the length $T$ of the training video clip is set to 2. We freeze all batch normalization layers and minimize our loss using AdamW optimizer ($\beta = (0.9, 0.999)$, $eps = 10^{-8}$, and the weight decay is $10^{-4}$) with the initial learning rate $lr=10^{-4}$. The model is trained with batchsize 4 for 160 epochs on 4 TITAN RTX GPUs, taking about 1.5 days. In the inference stage, TransVOS with input resolution 480p only refers to the first and previous frames to segment the current frame. We conduct all inference experiments on a single TITAN RTX GPU.
	
	
	\subsection{Comparison with the State-of-the-art}
	\paragraph{DAVIS.} We compare the proposed TransVOS with the state-of-the-art methods on DAVIS benchmark \cite{perazzi2016benchmark, pont20172017}. We also present the results trained with additional data from YouTube-VOS \cite{xu2018youtube}. The evaluation results on DAVIS16-val and DAVIS17-val are reported in Table \ref{table:1}. When adding YouTube-VOS for training, our method achieves state-of-the-art performance on DAVIS17-val ($J\&F 83.9\%$), outperforming the online-learning methods with a large margin and have higher performance than the matching-based methods such as STM, RMNet and CFBI.
	Specifically, TransVOS outperforms transformer-based SST with 1.4\% in $(J\&F)$ and surpasses JOINT with gap of 0.4\% in $J\&F$. When only using DAVIS for training, our model achieves better quantitative results than those methods with same configuration and even better than several methods like FEELVOS and AGAME which apply YouTube-VOS for training.
	On DAVIS16-val, TransVOS has comparable performance with state-of-the-art methods. 
	Compared to KMN, our model has the same $J\&F$ score with a higher $J$ score while a slightly lower $F$ score. Since DAVIS 2016 is a single object dataset, segmentation details such as boundaries play an important role in performance evaluation. We believe that the Hide-and-Seek training strategy, which provides more precise boundaries, helps KMN a lot.
	We also report the results on the DAVIS17 test-dev in Table \ref{table:2}. Our TransVOS outperforms all the online-learning methods. Except slightly lower than KMN of $0.3\%(J\&F)$, TransVOS surpasses all the methods in the second part.
	
	\paragraph{YouTube-VOS.} Table \ref{table:2} shows comparison with state-of-the-art methods on YouTube-VOS 2018 validation \cite{xu2018youtube}. On this benchmark, our method obtains an overall score of 81.8\% and also outperforms all the methods in the first and second parts, which demonstrates that TransVOS is robust and effective. Specifically, the proposed TransVOS surpasses STM by 2.4\% in overall score. Note that we only refer to the first and previous frames to segment the current frame, while STM contains a large memory bank which saves a new memory frame every five frames. 
	Also, TransVOS outperforms KMN and CFBI with gaps of both 0.4\% in overall score. 
	Besides, it surpasses the most related transformer-based SST.
	

    \begin{table} [ht]
		\centering
		\resizebox{0.48\textwidth}{!}{
		\begin{tabular}{cccccc}
			\cmidrule(r){1-6}
			\multirow{2}{*}{Feature extractor} & $\mathcal{J}$ & $\mathcal{F}$ & $\mathcal{J\&F}$ & Parameters & \multirow{2}{*}{FPS} \\
			 & (\%) & (\%) & (\%) & (M) & \\
			\cmidrule(r){1-6}
			\textit{Siamese} & 64.9 & 72.4 & 68.6 & 35.61 & 18.1 \\
			\textit{Independent} & 72.8 & 80.3 & 76.5 & 43.08 & 17.0  \\
			\textit{Two-path} & 73.1 & 79.7 & 76.4 & 34.56 & 17.0  \\
			\cmidrule(r){1-6}
		\end{tabular}}
		\caption{Impacts of different types of feature extractors. Models are tested with the input resolution of 240p on DAVIS17-val.}
		\label{AB:6}
	\end{table}
	
	\begin{table} [t]
		\centering
		\resizebox{0.48\textwidth}{!}{
		\begin{tabular}{cccccccc}
			\cmidrule(r){1-8}
			\multirow{2}{*}{Components} &
			$\mathcal{J}_{M}$ & $\mathcal{J}_{R}$ & $\mathcal{J}_{D}$ & $\mathcal{F}_{M}$ &
			$\mathcal{F}_{R}$ & $\mathcal{F}_{D}$ & $\mathcal{J\&F}$ \\
			& (\%) & (\%) & (\%) & (\%) & (\%) & (\%) & (\%) \\
			\cmidrule(r){1-8} 
			w/o TD & 71.7 & 83.4 & 5.2 & 78.7 & 89.7 & 7.5 & 75.2 \\
			w/ TD & 73.1 & 86.6 & 1.8 & 79.7 & 91.5 & 5.3 & 76.4 \\
			\cmidrule(r){1-8}
	    \end{tabular}}
		\caption{Ablation studies of different components with input resolution 240p on DAVIS 2017 validation set. `TD' denotes the transformer decoder.}
		\label{AB:2}
	\end{table}

	\subsection{Ablation Study}
	We conduct all the ablation experiments on DAVIS17 validation \cite{pont20172017}. The model used in this section does not do pre-training on synthetic videos and the input resolution is 240p unless specified. And we test the model with only the first and previous frames referred by default. Here we list the ablation studies about two-path feature extractor, mask utilization, reference sets, and transformer structure. The exploration of the backbone, input resolution and training strategy are presented in the supplementary materials.
	
	\paragraph{Two-path feature extractor.} In Table \ref{AB:6}, we compare the proposed two-path feature extractor with i) existing approach of using two independent encoders (as in STM \cite{oh2019video}; denoted \textit{`Independent'}), and ii) using a siamese encoder and concatenating the object mask to the reference frame features (as in AGAME\cite{johnander2019generative}; denoted as \textit{`Siamese'}). Results showed that our two-path feature extractor employs fewer parameters (about 20\% reduction) than i), but obtains higher performance (+7.8\% in $J\&F$ score) than ii).
	
	\paragraph{Mask utilization.} To demonstrate the effectiveness of our two-path feature extractor, we implement three typical ways to utilize the predicted masks of past frames. (1) the predicted masks are multiplied with the encoded features of RGB frame, denoted as `\textit{multiply}'; (2) the encoded features of RGB frame and the predicted mask are multiplied firstly and then added to the former, denoted as `\textit{residual}'; (3) the predicted masks and the RGB frame are fed into a two-path feature extractor, denoted as `\textit{two-path}'. As shown in Table \ref{AB:1}, compared to directly multiply the predicted mask with encoded features (line 1) and fusing mask with residual structure (line 2), our two-path feature extractor gains $15.1\%(J\&F)$ and $8.0\%(J\&F)$ improvement.
	
	\paragraph{Reference sets.} We test how reference sets affect the performance of our proposed model. We experiment with four types of reference set configurations: (1) Only the first frame with the ground-truth masks; (2) Only the previous frame with its predicted mask; (3) Both the first and previous frame with their masks; (4) The reference set is dynamically updated by appending new frames with the predicted masks every 5 frames. As Table \ref{AB:1} shows, even with two frames referred, our model could achieve superior performance. Interestingly, we find that updating the memory every 5 frames as STM \cite{oh2019video} for VOS may not be beneficial in all methods. Because low-quality segmentation results of historical frames may mislead subsequent mask prediction.
	
	\paragraph{Transformer structure.} We explore the effectiveness and necessity of the transformer decoder in Table \ref{AB:2}. It can be seen that equipping with transformer decoder, our model obtains $1.2\%(J\&F)$ improvement over removing it. Therefore, it's essential to employ the transformer's decoder. 

	\section{Conclusions}
	In this paper, we propose a novel transformer-based pipeline, termed TransVOS, for semi-supervised video object segmentation (VOS). Specifically, we employ the vision transformer to model the spatial and temporal relationships at the same time among reference sets and query frame. Moreover, we propose a two-path feature extractor to encode the reference sets and query frames, which dramatically slim the existing VOS framework while keeping the performance. Our TransVOS achieves the top performance on several benchmarks, which demonstrates its potential and effectiveness.
	
	\bibliography{aaai22}
	
	\clearpage
	\section{Appendix}
	\paragraph{Datasets and Evaluation Metrics}
	We evaluate our approach on DAVIS \cite{perazzi2016benchmark, pont20172017} and YouTube-VOS \cite{xu2018youtube} benchmarks. Both DAVIS2016 and DAVIS2017 have experimented. DAVIS2016 is an annotated single-object dataset containing 30 training video sequences and 20 validation video sequences. DAVIS2017 is a multi-objects dataset expanded from DAVIS2016, including 60 training video sequences, 30 validation video sequences, and 30 test video sequences. YouTube-VOS dataset is a large-scale dataset in VOS, having 3471 training videos and 474 validation videos. And each video contains a maximum of 12 objects. The validation set includes seen objects from 65 training categories and unseen objects from 26 categories, which is appropriate for evaluating algorithms' generalization performance. We use the evaluation metrics provided by the DAVIS benchmark to evaluate our model. $J\& F$ evaluates the general quality of the segmentation results, $J$ evaluates the mask $IoU$ and $F$ estimates the quality of contours.
	
	\paragraph{Backbone.} We experiment with different backbones, ResNet18 and ResNet50 \cite{he2016deep}. As shown in Table \ref{AB:3}, TransVOS with smaller backbone ResNet18 runs faster (7fps improvement) than ResNet50 while the performance drops $4.1\%J\&F$. 
	Therefore, we compare our TransVOS with ResNet50 as the backbone to other state-of-the-art methods.

	\begin{table} [ht]
		\centering
		\resizebox{0.48\textwidth}{!}{
			\begin{tabular}{ccccccccc}
				\cmidrule(r){1-9}
				\multirow{2}{*}{Backbone} & $\mathcal{J}_{M}$ & $\mathcal{J}_{R}$ & $\mathcal{J}_{D}$ & $\mathcal{F}_{M}$ & $\mathcal{F}_{R}$ & {$\mathcal{F}_{D}$} & $\mathcal{J\&F}$ & \multirow{2}{*}{FPS} \\
				& (\%) & (\%) & (\%) & (\%) & (\%) & (\%) & (\%) \\
				\cmidrule(r){1-9}
				ResNet18 & 68.8 & 80.6 & 7.7 & 75.9 & 86.3 & 9.1 & 72.3 & 24.0 \\
				ResNet50 & 73.1 & 86.6 & 1.8 & 79.7 & 91.5 & 5.3 & 76.4 & 17.0 \\
				\cmidrule(r){1-9}
		\end{tabular}}
		\caption{Ablation studies of different backbone with input resolution 240p on DAVIS 2017 validation set.}
		\label{AB:3}
	\end{table}
	
    \paragraph{Training strategy.} We conduct experiments to explore the effectiveness of pre-trainng on synthetic videos. As Table \ref{AB:4} shows, without pre-training, our model only drops by 1.5\% in $(J\&F)$, which means our proposed TransVOS can learn general and robust target object appearance even training with small dataset.
	
	\begin{table} [ht]
		\centering
		\resizebox{0.48\textwidth}{!}{
			\begin{tabular}{cccccccc}
				\cmidrule(r){1-8}
				\multirow{2}{*}{Training strategy} & $\mathcal{J}_{M}$ & $\mathcal{J}_{R}$ & $\mathcal{J}_{D}$ & $\mathcal{F}_{M}$ & $\mathcal{F}_{R}$ & {$\mathcal{F}_{D}$} & $\mathcal{J\&F}$ \\ 
				& (\%) & (\%) & (\%) & (\%) & (\%) & (\%) & (\%) \\
				\cmidrule(r){1-8}
				w/o pre-training & 73.1 & 86.6 & 1.8 & 79.7 & 91.5 & 5.3 & 76.4 \\
				w/ pre-training & 74.4 & 85.6 & 6.8 & 81.4 & 91.3 & 8.1 & 77.9 \\
				\cmidrule(r){1-8}
		\end{tabular}}
		\caption{Training data analysis on DAVIS 2017 validation set. We do abaltion studies to explore how the pre-training affects our model's performance.}
		\label{AB:4}
	\end{table}
	
	\paragraph{Input resolution.} We adjust the input resolution of the model as shown in Table \ref{AB:5}, from which we can see that our method achieves better performance with a larger input size. TransVOS with half input resolution runs faster (11.8fps improvement) while the performance drops $4.0\%J\&F$. Therefore, we compare our TransVOS with input resolution 480p to other state-of-the-art methods.
	
	\begin{table} [ht]
		\centering
		\resizebox{0.48\textwidth}{!}{
		\begin{tabular}{ccccccccc}
			\cmidrule(r){1-9}
			Input & $\mathcal{J}_{M}$ & $\mathcal{J}_{R}$ & $\mathcal{J}_{D}$ & 
			$\mathcal{F}_{M}$ & $\mathcal{F}_{R}$ & $\mathcal{F}_{D}$ & $\mathcal{J\&F}$ &
			\multirow{2}{*}{FPS} \\ 
			resolution & (\%) & (\%) & (\%) & (\%) & (\%) & (\%) & (\%) & \\
			\cmidrule(r){1-9}
			240p & 74.4 & 85.6 & 6.8 & 81.4 & 91.3 & 8.1 & 77.9 & 17.0 \\
			480p & 81.4 & 90.6 & 7 & 86.4 & 93.7 & 8.8 & 83.9 & 5.2 \\
			\cmidrule(r){1-9}
		\end{tabular}}
		\caption{Input resolution analysis. We compared models with different input resolution on DAVIS 2017 validation set.}
		\label{AB:5}
	\end{table}

\end{document}